\pdfoutput=1

\documentclass[11pt]{article}

\usepackage[final]{acl}

\usepackage{times}
\usepackage{latexsym}

\usepackage[T1]{fontenc}

\usepackage[utf8]{inputenc}

\usepackage{microtype}

\usepackage{inconsolata}
\usepackage{url}
\usepackage{graphicx}
\usepackage{booktabs}
\usepackage{amsmath} 
\usepackage{tcolorbox}
\usepackage{tikz}
\usepackage{multirow}
\usepackage{xspace}

\definecolor{c0}{cmyk}{1,0.3968,0,0.2588} 

\newtcbox{\pattern}{on line,colback=c0!10,colframe=white,size=fbox,arc=3pt, box align=base,before upper=\strut,
top=-2pt, bottom=-2pt, boxrule=0pt}
\title{Teaching Language Models to Self-Improve \\ 
by Learning from Language Feedback}

\author{Chi Hu$^1$ Yimin Hu$^1$ Hang Cao$^1$ 
{\bf Tong Xiao$^{1,2}$\thanks{ Corresponding author.} Jingbo Zhu$^{1,2}$} \\
	\textsuperscript{1}NLP Lab, School of Computer Science and Engineering, \\ Northeastern University, Shenyang, China\\
	\textsuperscript{2}NiuTrans Research, Shenyang, China\\
	\ttfamily{huchinlp@gmail.com}\\
	\ttfamily{\{xiaotong,zhujingbo\}@mail.neu.edu.cn}
}

\date{}

\begin{document}
\maketitle
\begin{abstract}
Aligning Large Language Models (LLMs) with human intentions and values is crucial yet challenging. Current methods primarily rely on human preferences, which are costly and insufficient in capturing nuanced feedback expressed in natural language. In this paper, we present Self-Refinement Tuning (SRT), a method that leverages model feedback for alignment, thereby reducing reliance on human annotations. SRT uses a base language model (e.g., Tulu2) to generate initial responses, which are critiqued and refined by a more advanced model (e.g., GPT-4-Turbo). This process enables the base model to self-evaluate and improve its outputs, facilitating continuous learning. SRT further optimizes the model by learning from its self-generated feedback and refinements, creating a feedback loop that promotes model improvement. Our empirical evaluations demonstrate that SRT significantly outperforms strong baselines across diverse tasks and model sizes. When applied to a 70B parameter model, SRT increases the win rate from 9.6\% to 25.8\% on the AlpacaEval 2.0 benchmark, surpassing well-established systems such as GPT-4-0314, Claude 2, and Gemini. Our analysis highlights the crucial role of language feedback in the success of SRT, suggesting potential for further exploration in this direction.
\end{abstract}

\section{Introduction}
Recent advances in Large Language Models (LLMs) have revolutionized the field of natural language processing. These models have demonstrated remarkable capabilities in various tasks such as open-ended generation, question answering, and mathematical reasoning \cite{Brown2020LanguageMA, Chowdhery2022PaLMSL, Ouyang2022TrainingLM, Bubeck2023SparksOA}. However, despite their impressive performance, LLMs occasionally generate content that can be untruthful or harmful. This highlights the need to align language models with human intentions and values to ensure safe and controllable deployment \cite{Weidinger2021EthicalAS, Bai2022ConstitutionalAH, Perez2022RedTL}.

\begin{figure}[t]
    \centering
    \includegraphics[scale=0.25]{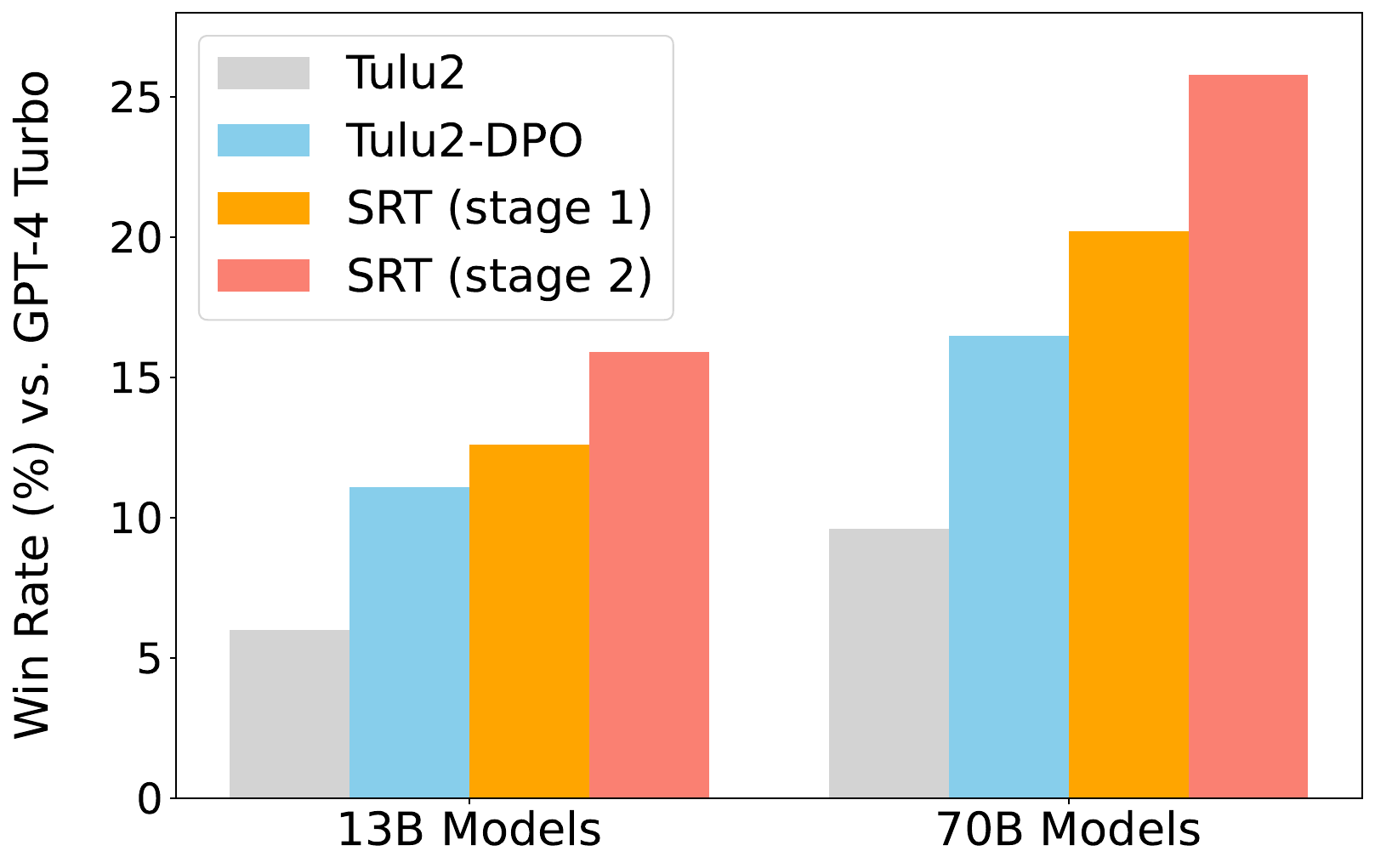}
    \caption{Results on AlpacaEval 2.0. SRT significantly boosts the performance of the base Tulu2 models. We report the win rates against GPT-4 Turbo.}
    \label{fig:alpaca_eval_2_short}
\end{figure}

Many current methods to align LLMs, such as Reinforcement Learning from Human Feedback (RLHF) and Direct Preference Optimization (DPO), rely on human preferences \cite{Christiano2017DeepRL, Jaques2019WayOB, Stiennon2020LearningTS, Rafailov2023DirectPO}. These preferences are typically expressed through rankings or scores. However, there are two main challenges with these methods. Firstly, annotating human preferences is expensive and time-consuming, which makes it difficult to scale these techniques. For instance, \citet{Scheurer2023TrainingLM} spent \$40k to annotate approximately 60K feedback instances for a single summarization task. Secondly, simple rankings or scores may not fully capture the subtlety and complexity of human preferences. This can limit the depth of feedback provided to models for improvement \cite{Myers2021LearningMR, casper2023open}. Humans, on the other hand, can derive insights from minimal language feedback, indicating that there is potential for more sophisticated and efficient alignment methods \cite{Scheurer2022TrainingLM, Chen2023ImprovingCG}.

In response to these challenges, we introduce a new method for aligning language models, which we call Self-Refinement Tuning (SRT). SRT obviates the need for human preferences with a two-stage learning process. In the first stage, SRT employs a powerful model like GPT-4 to critique and refine the outputs from a base model, such as Tulu2 \cite{Ivison2023CamelsIA}. The base model is then fine-tuned on critiques and refinements, enabling self-evaluation and improvement. In the second stage, SRT further boosts the model by learning from self-feedback. Specifically, SRT uses the fine-tuned model to generate model preferences, i.e., pairs of outputs and refinements. Subsequently, SRT optimizes the model on these preferences using DPO \cite{Rafailov2023DirectPO}.

\begin{figure*}
	\centering
	\includegraphics[width=160mm]{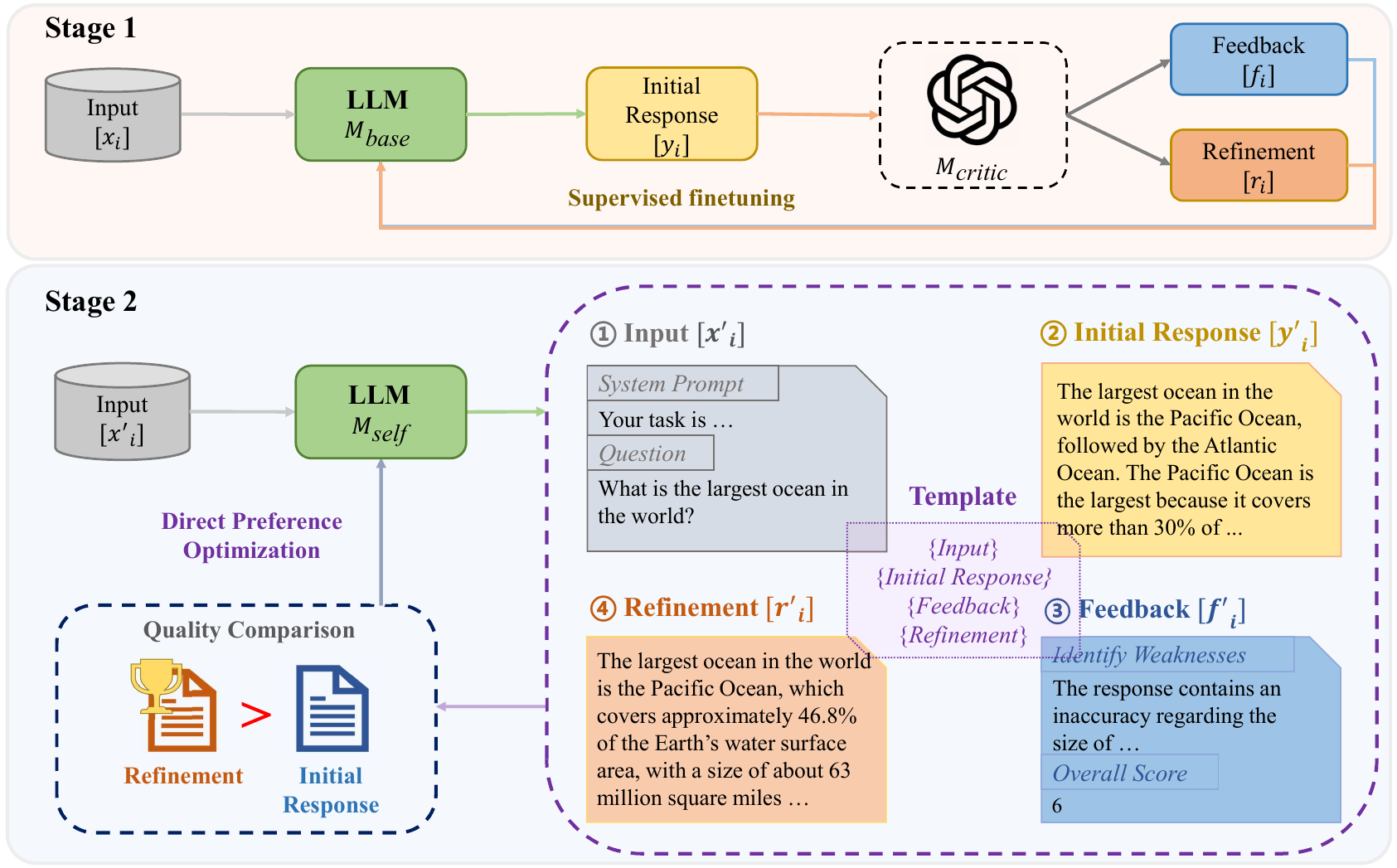}
	\caption{An overview of Self-Refinement Tuning (SRT). In the first stage (above), SRT teaches the base model to self-improve by fine-tuning it on the feedback and refinements from a powerful critic model. In the second stage (bottom), SRT enables the model to learn from its self-generated feedback and refinements.}
 \label{fig:main_method}
\end{figure*}

We evaluate SRT on open-ended generation, question answering, and mathematical reasoning. Empirical results show that SRT consistently outperforms baseline models of sizes from 7B to 70B, with an average performance enhancement of 3.7 to 4.0 points. These improvements are obtained using merely 22K feedback instances annotated by GPT-4 Turbo. Figure \ref{fig:alpaca_eval_2_short} shows that the SRT significantly improves Tulu2 models using model-generated feedback. The strongest model trained with SRT attains a 25.8\% win rate against GPT-4 Turbo on the AlpacaEval 2.0 benchmark. This performance surpasses established systems such as GPT-4 0314, Mistral Medium, Claude, and Gemini. Our analysis confirms that the success of SRT primarily stems from its language feedback feature, which identifies weak areas and offers valuable suggestions for improvement.

\section{Related Work}
This section briefly reviews two critical areas in the alignment of LLMs: learning from AI feedback and learning to self-improve. Our work intersects with these domains and addresses existing challenges.

\paragraph{Learning from AI Feedback.} Learning from human feedback is the key to the success of state-of-the-art language models such as GPT-4 \cite{openai2023gpt4} and Gemini \cite{geminiteam2023gemini}. However, acquiring high-quality human feedback is both costly and time-consuming \cite{Christiano2017DeepRL, Jaques2019WayOB, Stiennon2020LearningTS, Rafailov2023DirectPO}. This has led to a growing interest in harnessing AI-generated feedback to enhance LLMs \cite{Lee2023RLAIFSR, Roit2023FactuallyCS, hu2024rankprompt}. For instance, \citet{hu2024rankprompt} employs LLMs to annotate ranking-based preferences. Our research diverges from this approach by utilizing a more comprehensive range of feedback, encompassing identified weaknesses and proposed improvements. A similar approach is Reinforcement Learning from AI feedback (RLAIF, \citealp{Bai2022ConstitutionalAH}), which uses LLMs to generate critiques and refinements. However, RLAIF encounters challenges such as computational inefficiency and unstable training dynamics due to the inherent complexities of reinforcement learning techniques \cite{casper2023open, Shen2023LargeLM}. We address these challenges by unifying the generation of feedback and refinement into instruction-following, thereby simplifying the training process.

\paragraph{Learning to Self-Improve.} Using Large Language Models (LLMs) to identify and correct their own errors has become increasingly popular. This self-improvement capability has significantly enhanced performance across various tasks, such as question-answering, code generation, and mathematical reasoning \cite{Shinn2023ReflexionLA, Madaan2023SelfRefineIR, Chen2023TeachingLL}. \citet{Pan2023AutomaticallyCL} conducted a comprehensive survey of this field. However, these approaches rely on critique and refinement skills that are generally lacking in open-source language models \cite{Valmeekam2023CanLL, Huang2023LargeLM}. This underscores the urgent need to train open-source models for self-improvement. Unlike previous methods that train separate models for generation, critique, and refinement \cite{Yasunaga2020GraphbasedSP, Welleck2022GeneratingSB}, our approach employs a single model for all tasks, facilitating knowledge transfer across them. Additionally, our method is designed for the general alignment of LLMs, in contrast to prior methods that are task-specific \cite{Wang2023LearningFM, Yu2023TeachingLM, Lu2023SELFSW}.

\section{Methodology}
Figure \ref{fig:main_method} presents the two-stage Self-Refinement Tuning (SRT) process. SRT aims to enhance the capabilities of a base language model (denoted as $M_{base}$) by harnessing learning from AI feedback (LAIF), thereby reducing the reliance on human feedback. In the first stage, we train $M_{base}$ to self-improve by learning from feedback and refinements annotated by more powerful models. In the second stage, we further optimize the model using its self-generated feedback.

\subsection{Training Models for Self-Improvement}
\subsubsection{Collecting Feedback and Refinements}
In the first stage, we collect the training data for self-improvement from the interaction of two models. Specifically, $M_{base}$ interacts with a stronger model, denoted as $M_{critic}$. Given a set of instructions $\textbf{x} = [x_1, ..., x_N]$, $M_{base}$ generates initial responses $\textbf{y} = [y_1, ..., y_N]$. $M_{critic}$ then provides feedback $\textbf{f} = [f_1, ..., f_N]$ on these responses. As delineated in Table \ref{table:feedback_template}, we instruct $M_{critic}$ to generate 1) analyses of the weaknesses, 2) overall quality scores ranging from 1 to 10, and 3) suggestions for enhancing the responses. Subsequently, $M_{critic}$ generates the refinements (i.e., improved responses) $\textbf{r} = [r_1, ..., r_N]$ according to its previous feedback. 

\begin{table}[t]
\small
\begin{tcolorbox}[colback=gray!5]
\textbf{\#\#\# Instruction}

\textcolor[rgb]{0,0,0.7} {\{Instruction\}}

\textbf{\#\#\# Response}

\textcolor[rgb]{0,0,0.7} {\{Response\}}

\textbf{\#\#\# Requirements}

As a AI assistant instructor, your task is to provide constructive feedback on responses generated by the assistant.
Ensure that your feedback focuses on the original instruction and do not introduce new requirements.
Follow these steps to ensure your feedback is effective and beneficial:

1. Identify Weaknesses: Carefully review the response and pinpoint any areas where it falls short.

2. Offer Actionable Advice: Provide specific suggestions on how the response can be improved.

3. Conclude with a Rating: After providing feedback, score the response on a scale from 1 to 10, with 1 being the lowest and 10 the highest. Use the format: "Overall Score: [[1-10]]".

4. Show an Improved Response: Offer an improved version that incorporates your feedback. Clearly indicate the enhanced response with the heading: "\#\#\# Improved Response".

\textbf{\#\#\# Feedback}
\end{tcolorbox}
\caption{The template we used for obtaining feedback from LLMs. It instructs LLMs to evaluate and refine the response to a given instruction.}
\label{table:feedback_template}
\end{table}

The critique-refinement process can be iterative: after the first iteration, we obtain a set of instructions, initial outputs, critiques, and refinements. We then use the refinements as the inputs for the next iteration and continue the process until the quality of the outputs no longer improves. Intriguingly, we find that a \textit{single iteration} of refinement is typically adequate, with additional iterations contributing only marginal improvements. See Section \ref{sec:details} for more details of the process of collecting the feedback and refinement.

\subsubsection{Self-Improvement Training}
After collecting the feedback and refinements, we fine-tune $M_{base}$ to identify and rectify its own errors or undesirable behaviors, thereby improving its performance. To facilitate this, we reformat the collected data into sequences of sentences as follow:
\pattern{\small $Instruction$ $\rightarrow$ \small $Response$ $\rightarrow$ \small $Feedback$ $\rightarrow$ $Refinement$}
By doing so, we can train the model using a language modeling objective. This method is somewhat akin to Chain of Hindsight \cite{Liu2023ChainOH}, where the model is trained to produce outputs from provided instructions and feedback. However, our approach differs in training the model to generate the feedback and refinements sequentially instead of merely predicting the outputs. More formally, we want to optimize:
\begin{equation}
\mathcal{L} = -\frac{1}{N}\sum_{i=1} \log P(y_i, f_i, r_i | x_i)
\end{equation}
where $N$ is data size, $x_i$ is the input instruction, $y_i$ is the initial output, $f_i$ is the feedback, and $r_i$ is the best refinement with the highest score. The input instruction is masked from the loss calculation during the training phase. This results in a self-improvement model, denoted by $M_{self}$, which can provide feedback on its outputs and refine them until they are satisfactory. In Section \ref{sec:experiments}, we show that $M_{self}$ significantly outperforms $M_{base}$.

\subsection{Scaling SRT through Self-Feedback}
We now describe how $M_{self}$ can be used to scale the learning from AI feedback. Specifically, we leverage $M_{self}$ to generate preference data and for further model optimization. Our approach is similar to Self-Rewarding \cite{Yuan2024SelfRewardingLM} but deviates from the strategy for acquiring superior responses. Rather than sampling multiple responses and selecting the optimal one, we directly generate an improved response via self-refinement, thereby enhancing efficiency.

More concretely, we use model $M_{self}$ to generate initial responses $\textbf{y'} = [y'_1, ..., y'_N]$ from a distinct instruction set $\textbf{x'} = [x'_1, ..., x'_N]$. The model critiques and refines these responses into feedback $\textbf{f'}$ and refinements $\textbf{r'}$. We filter to ensure refinements exceed initial response quality, based on model-assigned scores. This yields preference pairs of initial responses and refinements, which can used for DPO training \cite{Rafailov2023DirectPO}.

\section{Experiments}
\subsection{Implementation Details}
\label{sec:details}
In this subsection, we outline the specifics of SRT training, which includes the choice of base models, the procedure for feedback annotation and cleaning, and the comprehensive model training process.

\paragraph{Models.}
To evaluate the efficacy of SRT, we employ three fine-tuned LLaMA2 models as our base models. These include Tulu2-7B, Tulu2-13B, and Tulu2-70B, all trained on the Tulu-2 Mixture \cite{Ivison2023CamelsIA}. Given that these datasets are partially distilled from models such as GPT-4, they can also be referred to as \textbf{dSFT}, an acronym for distilled Supervised Fine-Tuning \cite{Tunstall2023ZephyrDD}. In the first stage of SRT, we use Tulu2-7B to generate initial responses and employ GPT-4 Turbo (gpt-4-1106-preview) to generate language feedback and refinements. The GPT-4 critic exhibits a 78.9\% agreement rate with human preferences on the HH-RLHF dataset \cite{bai2022training}. We then fine-tune all base models on these refinements. In the second stage, these fine-tuned models independently generate feedback and refinements for DPO training, which we refer to as \textbf{sDPO} (self-feedback DPO). To measure the effectiveness of self-feedback, we further compare our models with other DPO variants trained on the UltraFeedback \cite{Cui2023UltraFeedbackBL} dataset. These include Tulu2-DPO-7B, Tulu2-DPO-13B, and Tulu2-DPO-70B. We refer to these models as \textbf{dDPO}, an acronym for distilled DPO \cite{Tunstall2023ZephyrDD}.

\begin{figure}[t]
    \centering
    \includegraphics[scale=0.25]{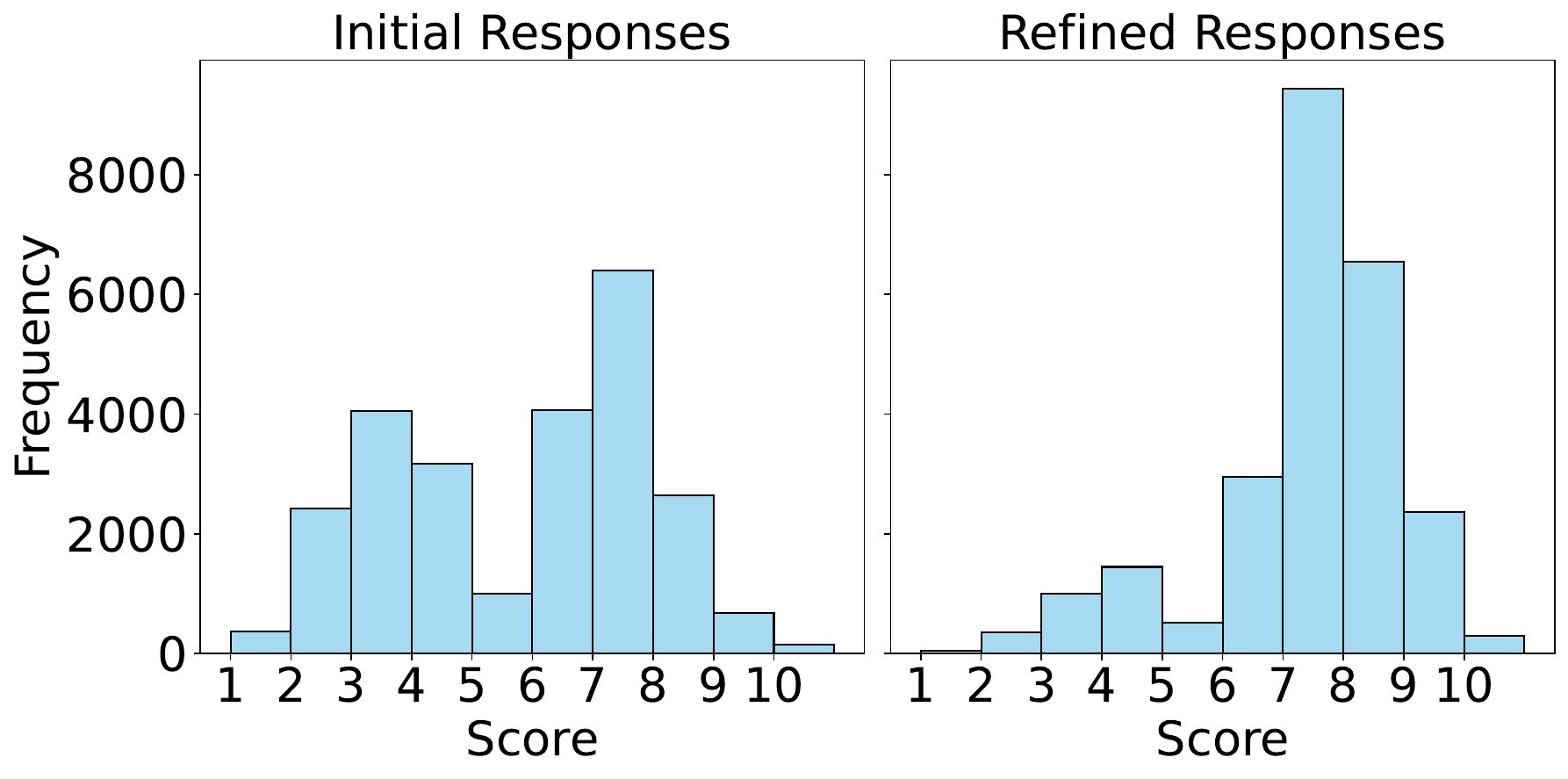}
    \caption{The score distribution of 25K initial responses (left) and refined responses (right). The initial responses are generated by Tulu2-7B and are then refined and scored by GPT-4 Turbo using the template presented in Table \ref{table:feedback_template}.}
    \label{fig:score_distribution}
\end{figure}

\paragraph{Details on Feedback and Refinements Annotation.}
In our two-stage process, we initially select 25,000 instructions from the Tulu-2 Mixture, followed by 75,000 instructions for the second stage. We adopt the approach of \citet{Li2023SelfAlignmentWI}, choosing examples of a single conversation turn to streamline the refinement process. For generating responses, we use a sampling temperature of 0.7, while for feedback and refinement generation, we set the temperature to 0 and restrict the maximum new tokens to 2,048. We use the prompt template shown in Table \ref{table:feedback_template} for annotating feedback and refinements. Figure \ref{fig:score_distribution} shows the score distribution of the initial outputs of Tulu2-7B and their subsequent refinements. On average, the refinements enhance the score by 1.5 points compared to the initial outputs. Table \ref{table:multi_iteration} shows one iteration of critique-and-refinement is sufficient.

\paragraph{Post-processing of Feedback and Refinements.}
Our goal is to compile instances that form a consistent sequence of instruction-response-feedback-refinement. However, even sophisticated models like GPT-4 Turbo occasionally fail to conform strictly to the format requirements. As a result, we apply the following filtering rules:

\textbf{Rule \#1:} The feedback should include potential weaknesses, overall scores, and suggestions\footnote{We identify these elements by looking for keywords such as `\#\#\#Feedback,' `\#\#\#Overall Score,' and `\#\#\#Improved Response.'}.

\textbf{Rule \#2:} The quality of the refinement should not be lower than that of the initial response.

Note that we evaluate the initial responses and their subsequent refinements independently. This is important as the critic sometimes gives higher scores when it has been conditioned with prior feedback, regardless of the quality of the refinement. After applying these filters, we obtain 22K and 63K valid feedback and refinements for the initial and subsequent stages, respectively. 

\paragraph{Training Details.}
\label{sec:training_details}
The base models are trained using the Tulu-2 Mixture dataset, following the same settings as \citet{Ivison2023CamelsIA}. In the first stage of SRT, we fine-tune the base models for five epochs. We concatenate each instruction's initial response, feedback, and refinement into sequences. Following \citet{Ivison2023CamelsIA}, we set the maximum sequence length to 8,192 but use a smaller global batch size of 32. We use a cosine learning rate scheduler with a peak learning rate of 1e-5 and 10\% warmup steps. In the second stage, we train our DPO models for one epoch with a $\beta$ value of 0.01. We maintain the same optimizer and learning rate scheduler as the first stage but adjust the maximum learning rate to 5e-7 to ensure training stability. We use the HuggingFace TRL library\footnote{\url{https://github.com/huggingface/trl}} to train our models, and use FlashAttention-2 \cite{Dao2023FlashAttention2FA} and DeepSpeed Zero-3 \cite{Rajbhandari2019ZeROMO} to speedup training.

\begin{table}[t!]
\centering
\footnotesize
\begin{tabular}{lcc}
\toprule
\textbf{Iteration} & \textbf{Average Score} \\
\midrule
Initial Responses & 5.42 \\
\midrule
Iteration 1 & 6.93 \\
Iteration 2 & 7.04 \\
Iteration 3 & 6.91 \\
Iteration 4 & 6.85 \\
\bottomrule
\end{tabular}
\caption{The average score of the refinements generated by GPT-4 Turbo at different iterations. The results are obtained from 1,000 samples from the Tulu-2 Mixture dataset. The initial responses are generated by Tulu2-7B.}
  \label{table:multi_iteration}
\end{table}
\begin{table*}[t]
\centering
\resizebox{\textwidth}{!}{
\begin{tabular}{lr|cccc|r}
\toprule
\multirow{2}{*}{\textbf{Model}} & \multirow{2}{*}{\textbf{Type}} & \textbf{GSM8k} & \textbf{BBH}  & \textbf{TydiQA} & \textbf{AlpacaEval} & \textbf{Average} \\
& & 8-shot CoT, EM & 3-shot CoT, EM & 1-shot, F1 & \% Win & - \\
\midrule
Tulu2 7B & dSFT & 34.0 & 48.5 & 46.4 & 73.9 & 50.7 \\
Tulu2-DPO 7B & dDPO & 34.5 & 45.5 & 44.5 & 85.1 & 52.4 \\
SRT 7B (\textit{stage 1}) & dSFT & \textbf{35.7} & \textbf{49.2} & \textbf{47.9} & 84.6 & \textbf{54.4} \\
SRT 7B (\textit{stage 2}) & sDPO & 34.2 & 47.3 & 47.2 & \textbf{85.3} & 53.4 \\
\midrule
Tulu2 13B & dSFT & 46.0 & 49.5 & 53.2 & 78.9 & 56.9 \\
Tulu2-DPO 13B & dDPO & 49.5 & 49.4 & 39.7 & 89.5 & 57.0 \\
SRT 13B (\textit{stage 1}) & dSFT & 48.2 & 50.4 & 56.4 & 88.7 & 60.9 \\
SRT 13B (\textit{stage 2}) & sDPO & \textbf{49.7} & \textbf{50.9} & \textbf{57.9} & \textbf{91.6} & \textbf{62.5} \\
\midrule
Tulu2 70B & dSFT & 73.0 & 68.4 & 53.6 & 86.6 & 70.4 \\
Tulu2-DPO 70B & dDPO & 71.5 & 66.0 & 35.8 & 95.1 & 67.1 \\
SRT 70B (\textit{stage 1}) & dSFT & 72.3 & \textbf{70.2} & 60.9 & 93.1 & 74.1 \\
SRT 70B (\textit{stage 2}) & sDPO & \textbf{73.9} & 69.7 & \textbf{61.8} & \textbf{95.2} & \textbf{75.1} \\
\bottomrule
\end{tabular}
}
\caption{Comparisons of the performance on four benchmarks. All models are fine-tuned from LLaMA2 pre-trained models. The term \textbf{dSFT} stands for distilled supervised fine-tuning, \textbf{dDPO} represents direct preference optimization with distilled feedback, and \textbf{sDPO} signifies direct preference optimization with self-generated feedback. The table presents the average scores for each benchmark, with the highest performing models of equal size highlighted in \textbf{bold}. The AlpacaEval results are win rates compared against GPT-3.5.}
\label{table:main_results}
\end{table*}
\paragraph{Evaluation Details.}
\label{sec:eval_setting}
We evaluate SRT across open-ended generation, reasoning, and question-answering. For open-ended generation, we test SRT on the AlpacaEval benchmark \cite{alpaca_eval}, which consists of 805 instructions. We report the win rate of our models compared to GPT-3.5 (text-davinci-003). Since this baseline may be outdated, we also compare our models with GPT-4 Turbo (gpt-4-1106-preview). We use the default configuration provided in the official library\footnote{We use `alpaca\_eval\_gpt4' as the judge for comparing GPT-3.5 and use `weighted\_alpaca\_eval\_gpt4\_turbo' for GPT-4 Turbo.}. For reasoning, we evaluate our models on GSM8K \cite{cobbe2021training} and Big-Bench-Hard (BBH, \citealp{suzgun2022challenging}). We report the exact match score (EM) on the test sets, using few-shot chain-of-thought prompting strategies identical to those used by \citet{Ivison2023CamelsIA}. For question-answering, we test with TydiQA \cite{clark2020tydi}, a multilingual benchmark that covers 11 different languages. In line with \citet{Ivison2023CamelsIA}, we report the F1 score under the Gold Passage (GP) setting, where the passage containing the answer is provided. To assess the ``self-evaluation" capability of SRT, we test our models on the HH-RLHF dataset \cite{bai2022training}, comparing model-predicted scores with human preferences. We evaluate using a sample of 500 data points from the HH-RLHF test set. Unless stated otherwise, we always select the refined responses as the outputs for the SRT models. We also limit the output refinement to one iteration, as additional iterations provide minimal benefits and significantly slow down the model's generation.

\subsection{Main Results}
\label{sec:experiments}
Table \ref{table:main_results} summarizes the experimental results on four benchmarks. The results indicate that SRT consistently improves model performance across various tasks and model sizes.
In the first stage, models trained with language feedback from GPT-4 significantly outperform the Tulu2 baselines. On average, the first stage of SRT enhances the performance by a margin of 3.7 to 4.0 across different model scales. Models trained during this stage are comparable to the Tulu2-DPO baselines on AlpacaEval but use considerably fewer feedback instances (22K vs 64K). These findings underscore the efficacy of training models to assess and refine their outputs by learning from the feedback of more advanced models.

In the second stage, SRT further boosts performance by scaling feedback with self-generated responses and refinements. The most notable improvement is observed on the AlpacaEval task. SRT's second stage is more effective with larger models (13B and 70B) but less with smaller ones (7B), which see an average drop of 1.0 points. This suggests that the inherent capabilities of a model may restrict its ability to learn from self-feedback.

Additionally, the enhancements observed in reasoning tasks, including GSM8K and BBH, are relatively slight, especially for the 7B and 13B models. This could be attributed to the limited reasoning capabilities present in these models.

\begin{table}[t!]
\centering
\footnotesize
\begin{tabular}{lrr}
\toprule
\textbf{Model} & \textbf{Win Rate (\%)} & \textbf{Length} \\
\midrule
GPT-4 0314 & 22.1 & 1371 \\
Mistral Medium & 21.9 & 1500 \\
Claude 2 & 17.2 & 1069 \\
Gemini Pro & 16.9 & 1315 \\
\midrule
Tulu2 13B & 6.0 & 993 \\
Tulu2-DPO 13B & 11.1 & 1440 \\
SRT 13B (\textit{stage 1}) & 12.6 & 1307 \\
SRT 13B (\textit{stage 2}) & 15.9 & 1285 \\
\midrule
Tulu2 70B & 9.6 & 960 \\
Tulu2-DPO 70B & 16.5 & 1480 \\
SRT 70B (\textit{stage 1}) & 20.2 & 1536 \\
SRT 70B (\textit{stage 2}) & \textbf{25.8} & 1663 \\
\bottomrule
\end{tabular}
\caption{AlpacaEval 2.0 results. We report the win rates compared to GPT-4 Turbo and the output length.}
  \label{table:alpaca_eval2_results}
\end{table}

\subsection{Results on AlpacaEval 2.0}
GPT-3.5 may be a relatively weak baseline for evaluating our strongest models. For example, both SRT 70B and Tulu2-DPO 70B models achieve win rates over 95\%. Therefore, we further evaluate these models on the more challenging AlpacaEval 2.0 benchmark, which employs GPT-4 Turbo as the baseline for calculating win rates. Table \ref{table:alpaca_eval2_results} presents the results of 13B and 70B models and several proprietary models. We can see that SRT models significantly outperform Tulu2 models. Notably, our strongest model, SRT 70B (stage 2), achieves a win rate of 25.8\%, outperforming Tulu2-DPO 70B by a substantial margin of 9.0\%. The model surpasses well-established models such as GPT-4 0314, Mistral Medium, Claude 2, and Gemini Pro. We also find that SRT encourages models to produce more verbose outputs, which could influence the observed performance improvement. However, despite generating shorter responses, our SRT 13B models achieve superior results compared to Tulu2-DPO models.

\begin{table}[t!]
\centering
\footnotesize
\begin{tabular}{lcc}
\toprule
\textbf{Model} & \textbf{Human Agreement (\%)} \\
\midrule
GPT-4 Turbo & 78.9 \\
\midrule
SRT 7B (\textit{stage 1}) & 74.4  \\
SRT 7B (\textit{stage 2}) & 68.2  \\
SRT 13B (\textit{stage 1}) & 72.2 \\
SRT 13B (\textit{stage 2}) & 68.9 \\
SRT 70B (\textit{stage 1}) & 72.8 \\
SRT 70B (\textit{stage 2}) & 72.3 \\
\bottomrule
\end{tabular}
\caption{Feedback accuracy of different models. We report the level of agreement with human preferences based on 500 samples from the HH-RLHF test set.}
  \label{table:feedback_accuracy}
\end{table}

\subsection{Results on HH-RLHF}
Table \ref{table:feedback_accuracy} presents a comparison of feedback accuracy among different models on the HH-RLHF test set. Feedback accuracy is evaluated based on the agreement of model outputs with human preferences. Specifically, for each pair of responses, the feedback is considered `correct' if the score of the `chosen' response exceeds that of the `rejected' response. Remarkably, GPT-4 Turbo surpasses all other models, achieving an agreement rate of 78.9\%. Intriguingly, during the initial stage of SRT, the 7B model outperforms larger models, reaching an agreement rate of 74.4\%. However, after the second-stage fine-tuning of SRT, the accuracy of the 7B and 13B models significantly declines, while the 70B models show slight changes.

\section{Analysis}
This section provides an in-depth examination of the factors that affect the performance of our methods. Initially, we evaluate the efficacy of self-refinement in SRT by contrasting it with the widely adopted re-ranking approach. Subsequently, we explore the influence of various elements in SRT, encompassing language feedback, the quality of refinements, and the quantity of training data. 
Through these analyses, we aim to deepen the understanding of our methods.

\begin{figure}[t]
    \centering
    \includegraphics[scale=0.25]{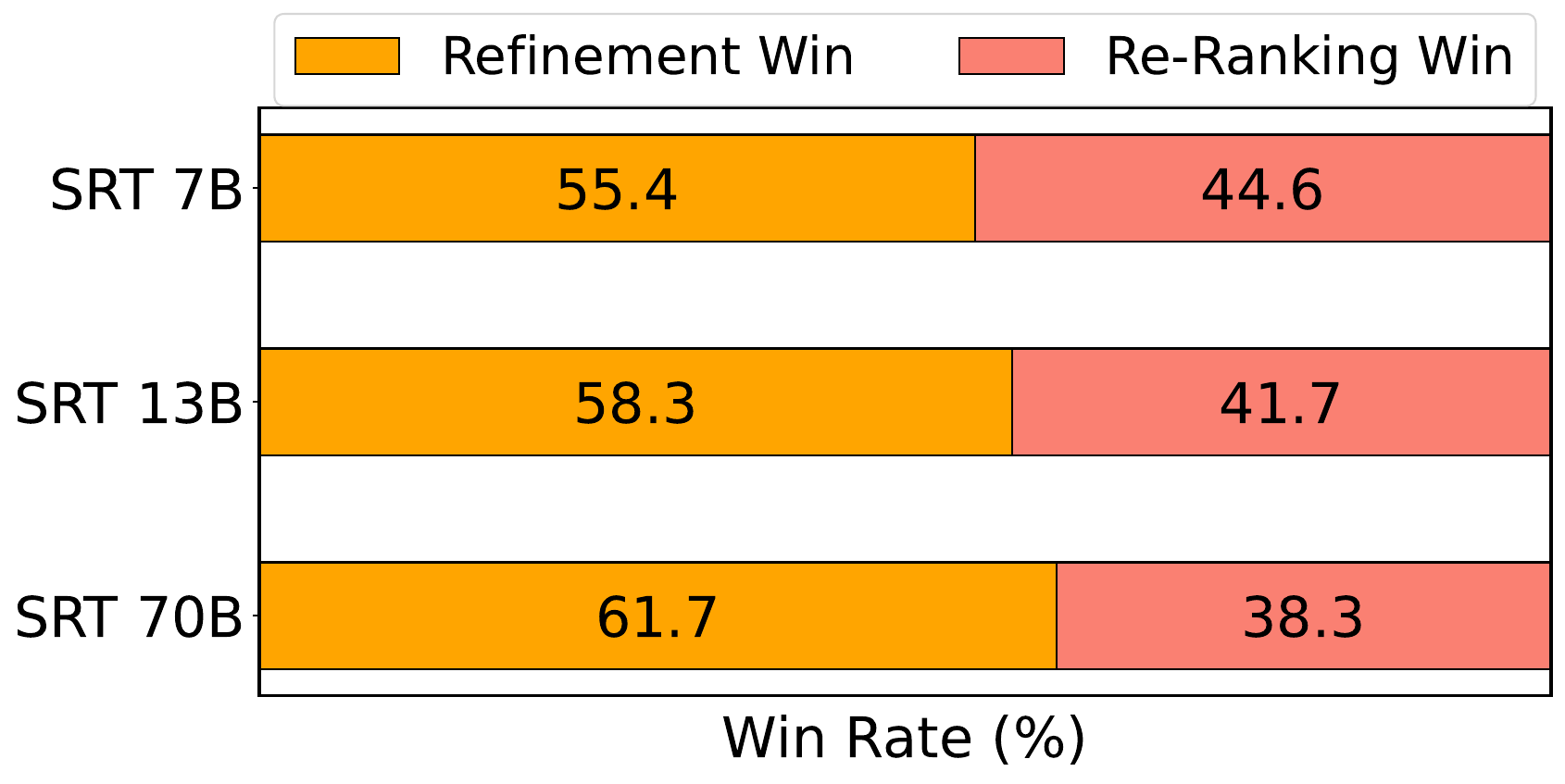}
    \caption{Self-Refinement vs. Re-Ranking over 16 candidates. The results are obtained on the AlpacaEval test set using models trained at the first stage of SRT.}
    \label{fig:re_ranking}
\end{figure}

\subsection{Self-Refinement vs. Re-Ranking}
Re-ranking is a widely adopted technique to bolster the performance of text generation systems. Given that our models can precisely assess the quality of their outputs, they can be effectively leveraged for re-ranking. In this study, we contrast the efficacy of self-refinement and re-ranking using the AlpacaEval test set. For self-refinement, we employ greedy decoding and refine the response once. For re-ranking, we sample 16 distinct responses using a temperature of 0.7. These responses are then re-ranked based on the scores predicted by the model itself. We then task GPT-4 for comparing the two sets of responses using the settings described in Section \ref{sec:eval_setting}. Figure \ref{fig:re_ranking} illustrates a direct comparison between self-refinement and re-ranking across 16 candidate responses. It is evident that self-refinement consistently surpasses re-ranking in performance across various models. This advantage becomes more pronounced with an increase in model size. Given the significant cost associated with generating multiple responses, self-refinement emerges as a more efficient and cost-effective approach for improving the performance of language models.

\begin{table}[t!]
\centering
\footnotesize
\begin{tabular}{lcc}
\toprule
\textbf{Model} & \textbf{Win Rate (\%)} & \textbf{Drop} \\
\midrule
SRT 13B (\textit{stage 1}) & 88.7 & - \\
\quad $-$ \textit{score}  & 87.6 & 1.1 \\
\quad $-$ \textit{suggestion}  & 85.9 & 2.8 \\
\quad $-$ \textit{weakness} & 85.5 & 3.2 \\
\quad $-$ \textit{all feedback}  & 83.6 & 5.1 \\
\midrule
Refinement Only & 84.2 & 4.5 \\
\bottomrule
\end{tabular}
\caption{Ablation study on the language feedback components. We report the win rates compared to GPT-3.5 using the same settings as Section \ref{sec:experiments}.}
  \label{table:language_feedback_ablation}
\end{table}

\subsection{Impact of Language Feedback}
A unique characteristic of SRT is its integration of language feedback, which offers insights into potential weaknesses, overall scores, and suggestions for improvement. Here, we ablate these components to investigate their influence on the final performance. We evaluate 13B models trained during SRT's first stage on AlpacaEval and compare these models with GPT-3.5. The results are presented in Table \ref{table:language_feedback_ablation}. We find that removing any part of the feedback leads to a drop in performance. Also, weaknesses and suggestions are almost equally important, while the score seems less necessary. Removing all feedback results in a significant performance drop of 5.1 points, which is even inferior to the performance achieved by training solely with refinement. To summarize, these findings emphasize the crucial role of language feedback in enhancing SRT's performance.

\begin{figure}[t]
    \centering 
    \includegraphics[scale=0.25]{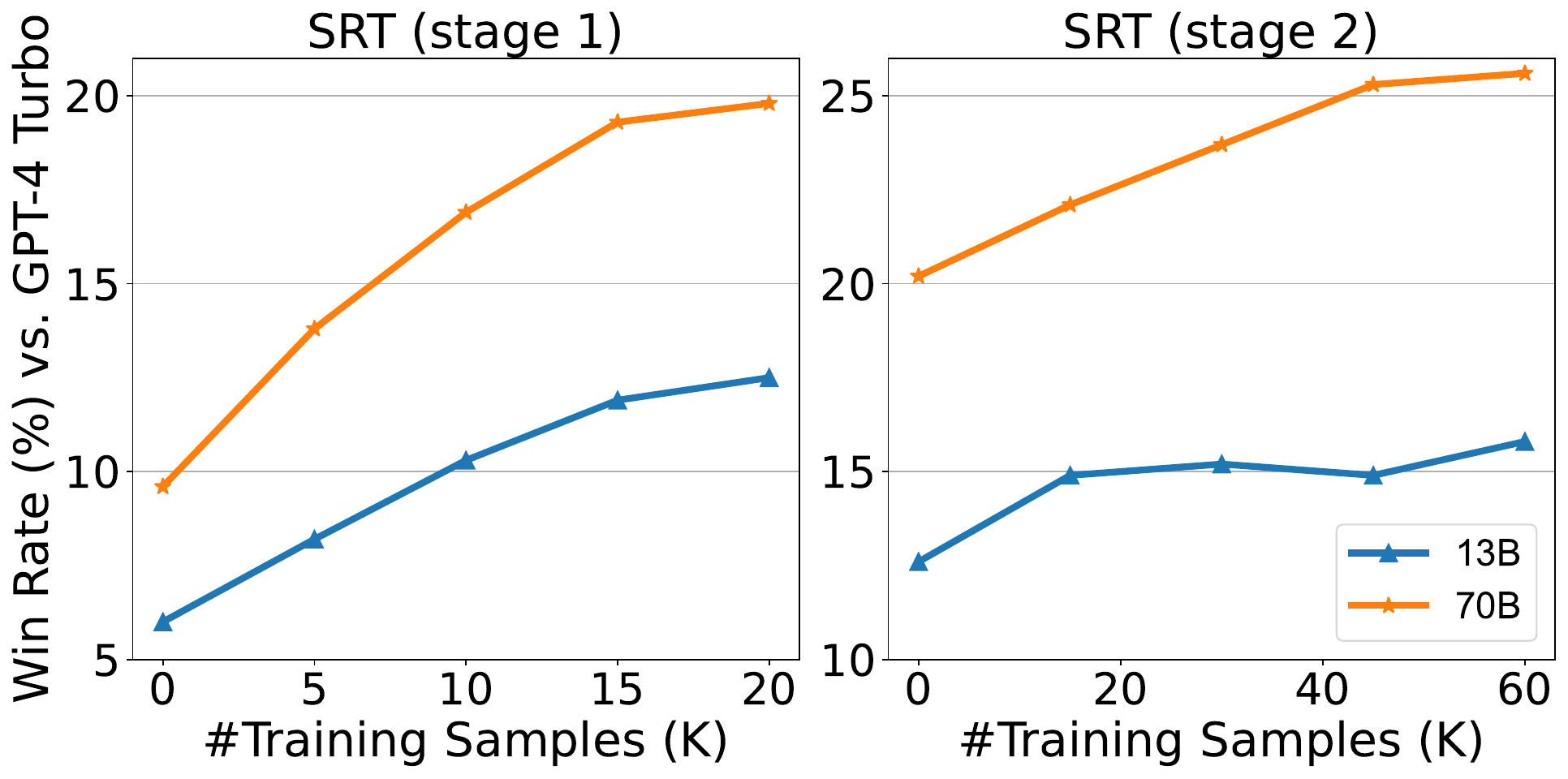}
    \caption{Win rates against GPT-4 Turbo by varying numbers of training samples for SRT models.}
    \label{fig:data_scaling}
\end{figure}
\subsection{Impact of Training Data Size}
In our primary experiments, we use 22K and 63K training samples for the first and second stages of SRT, respectively. We now delve further into the impact of data volume on the performance of SRT. We experiment with the challenging AlpacaEval 2.0 benchmark and replicate the training settings from Section \ref{sec:training_details}. Figure \ref{fig:data_scaling} illustrates the results of 13B and 70B models. As depicted, the performance almost monotonically increases with the variation in the number of training samples. Models of different sizes and stages exhibit diverse convergence speeds. Increasing the training data results in a more significant performance gain on the 70B models. However, we also find the 13B model's performance slightly deteriorates when the data volume increases from 36K to 48K. We hypothesize that this could be due to the presence of noise. To validate this hypothesis, we further analyze the impact of data quality on the performance of SRT.
\begin{figure}[t]
    \centering 
    \includegraphics[scale=0.25]{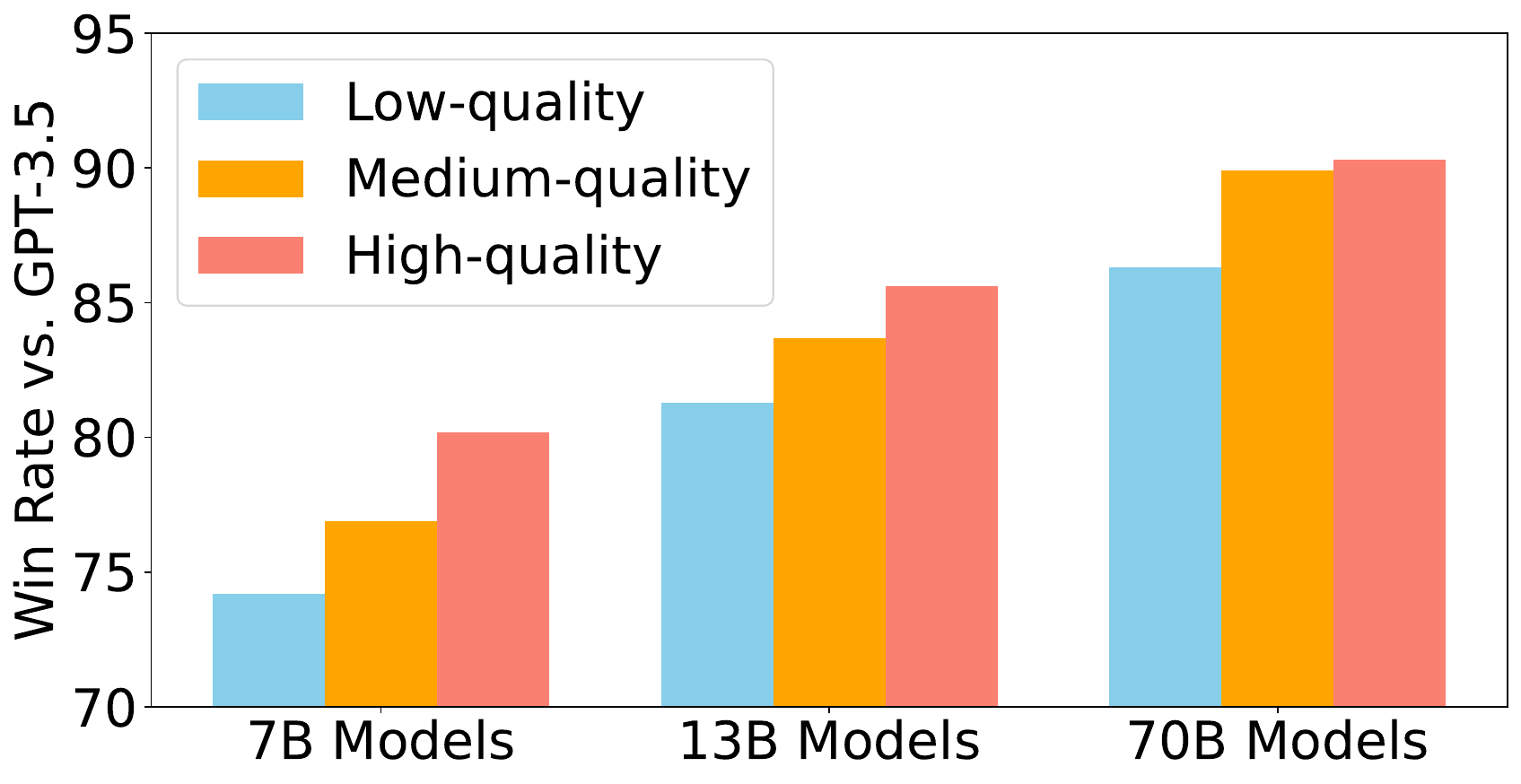}
    \caption{ AlpacaEval results (vs. GPT-3.5). The low-, medium-, and high-quality refinements have scores of 6, 7, and 8, respectively.}
    \label{fig:refinement_quality}
\end{figure}

\subsection{Impact of Refinement Quality}
\label{sec:refinement_quality}
In our primary experiment, we merely filter refinements that are of lower quality (reflected by overall scores) than the initial responses. Here, we delve deeper into this issue by conducting experiments with more fine-grained control over the refinement quality. To investigate this, we train models with varying levels of refinement quality and compare their performances. Specifically, we select samples of 2,000 refinements each from low-, medium-, and high-quality categories, corresponding to scores of 6, 7, and 8, respectively. As illustrated in Figure \ref{fig:refinement_quality}, there is a clear monotonic increase in the performance of SRT in line with the quality of refinements. The results hold with varying model sizes, suggesting that SRT can be further augmented by enhancing the refinement quality.


\section{Conclusion}
We have presented Self-Refinement Tuning (SRT) for aligning language models using language feedback. SRT initially teaches language models to assess and enhance their outputs by learning from feedback provided by more advanced models. Subsequently, SRT optimizes these models further by learning from their self-generated feedback. The primary benefits of SRT are twofold: (1) it obviates the need for human-annotated preferences, and (2) it obtains promising performance across a wide range of tasks and model sizes. In our analysis, we find that both high-quality language feedback and refinements are crucial for language models to learn to self-improve. Collectively, we demonstrate that SRT is an effective and efficient strategy for improving the performance of language models.

\section*{Acknowledgement}
This work was supported in part by the National Science Foundation of China (No.62276056), the Natural Science Foundation of Liaoning Province of China (2022-KF-16-01), the Fundamental Research Funds for the Central Universities (Nos. N2216016 and N2316002), the Yunnan Fundamental Research Projects (No. 202401BC070021), and the Program of Introducing Talents of Discipline to Universities, Plan 111 (No.B16009). The authors would like to thank anonymous reviewers for their insightful comments.

\section*{Limitations}
While SRT has demonstrated promising results across a variety of tasks, it is not without its limitations. A significant constraint lies in its dependency on a powerful critic model to provide feedback and refinements. Even state-of-the-art language models, such as GPT-4-Turbo, can sometimes generate feedback or refinements that are inaccurate or sub-optimal. For instance, the refined responses from GPT-4 yield an approximately average score of 7, indicating substantial room for improvement.

Moreover, one of the inherent limitation of self-refinement is it greatly increase the output length of language models. Although we find a single iteration of self-refinement is typically sufficient (in Table \ref{table:multi_iteration}), it still approximately doubles the output length of our models. The lengthy outputs lead to higher computational costs during decoding.

In the future, we will develop more accurate and efficient feedback mechanisms, optimize the self-refinement process to control output length, and explore ways to improve the quality of refinements.

\section*{Ethical Considerations}
SRT enhances language models using feedback and refinements produced by other models and the models themselves. This approach reduces dependency on human annotations, but it also introduces potential risks. Specifically, SRT can sometimes lead to unintended problems such as overfitting and reinforcing existing biases in the initial models. Therefore, it is crucial to develop more robust and trustworthy alignment methods to make sure that these language models are safe, fair, and controllable. Additionally, as the model improves, it may become increasingly complex and difficult to interpret. This can be problematic in applications where transparency and interpretability are essential. To address these issues, future works could include ethical guidance or external validation tools into the SRT process.
\bibliography{custom}



\end{document}